\documentclass[runningheads]{llncs}

\usepackage[T1]{fontenc}
\usepackage{graphicx}
\usepackage{amsmath}
\usepackage{amssymb}
\usepackage{array}
\usepackage{booktabs}
\usepackage{multirow}
\usepackage{xcolor}
\usepackage{pifont}
\usepackage{xspace}
\usepackage[hidelinks]{hyperref}

\newcommand{\Doe}{\ensuremath{\Delta_{\mathrm{oe}}}\xspace}
\newcommand{\Dor}{\ensuremath{\Delta_{\mathrm{or}}}\xspace}
\newcommand{\yes}{\ding{51}}
\newcommand{\no}{\ding{55}}

\begin{document}

\title{When Oracle Conditioning Misleads Deployment: Conditioning-Availability Bias in Echocardiographic Segmentation}
\titlerunning{Conditioning-Availability Bias in Segmentation}

\author{Dang P. M. Cao\inst{1,2}, Hieu D. Pham\inst{1}, Hieu Pham\inst{1,2,3}}

\authorrunning{D. P. M. Cao et al.}

\institute{
College of Engineering and Computer Science, VinUniversity, Hanoi, Vietnam
\and
VinUni-Illinois Smart Health Center, VinUniversity, Hanoi, Vietnam\\
\and
Center for Innovations in Health Sciences, VinUniversity, Hanoi, Vietnam\\
\email{24dang.cpm@vinuni.edu.vn, 24hieu.pd@vinuni.edu.vn}
}

\maketitle

\begin{abstract}
Conditional segmentation models may be trained and evaluated with auxiliary
signals cleaner than those available at deployment. We study this
protocol-level manifestation of shortcut learning and auxiliary-variable shift
in phase-conditioned echocardiographic segmentation. The complementary gap
pair $(\Doe,\Dor)$ measures loss on the deployable oracle--estimated pathway
and probes sensitivity on the oracle--random pathway. On held-out CAMUS data,
one strong-cyclic, oracle-selected run fails severely with estimated phase,
while sensitivity to incorrect phase persists across three runs. On
EchoNet-Dynamic, the current estimator remains usable, but random-phase testing
reveals strong latent sensitivity. Deployment-aware checkpoint selection and
phase perturbation reduce both gaps with little change in mean Dice.
Exploratory subgroup analyses quantify variation across measured strata, and a
downstream ejection-fraction (EF) audit shows that recovering segmentation does
not necessarily recover EF error or signed bias. Together, the gaps test
whether oracle-conditioned performance survives the inference pathway actually
available at deployment. Code: \href{https://github.com/minhdang050806/Oracle-Conditioning-Bias-in-Echocardiographic-Segmentation.git}{GitHub repository}.
\end{abstract}


\section{Introduction}

Segmentation systems increasingly use explicit conditioning or domain-specific
parameters, including FiLM features and scanner/protocol information
\cite{perez2018film,karani2018lifelong}. Benchmarks may supply these signals
exactly, whereas deployment receives an estimate, a noisy value, or none.
Oracle-only reporting can therefore overstate deployable performance.

We call this mismatch \emph{conditioning-availability bias}. It is a specific
evaluation-protocol form of shortcut learning and auxiliary-variable shift: the
benchmark supplies privileged conditioning, while deployment must use a less
reliable pathway \cite{geirhos2020shortcut}. The term identifies where the
mismatch enters the evidence chain; it is not intended as a separate theory of
bias.

The problem is primarily a deployment-validity failure and becomes
fairness-relevant only if conditioning availability or estimation quality
differs systematically across sites, scanners, operators, acquisition
conditions, or populations. We do not measure such access differences or
establish causal demographic unfairness.

We examine phase-conditioned segmentation in apical-four-chamber
echocardiography. Phase is derived from annotated end-diastolic (ED) and
end-systolic (ES) frames during training, but it must be estimated from the image
at inference. Comparing oracle, estimated, and deliberately incorrect phase
separates failure of the current deployment path from latent dependence on the
conditioning channel.

\noindent\textbf{Contributions.}
(1) We formalize a deployment audit through conditioning availability $r_A$,
estimation error $r_E$, and model sensitivity $r_S$, while stating which
components our data can and cannot measure. (2) We introduce the complementary
gaps $(\Doe,\Dor)$ to distinguish loss on the current deployable pathway from
latent sensitivity to incorrect conditioning. (3) Held-out experiments reveal
two patterns: seed-dependent estimated-path failure with replicated
random-phase sensitivity on CAMUS, and random-phase sensitivity despite a
usable estimated path on EchoNet-Dynamic. (4) Phase-head characterization and
cyclic-loss ablation separate estimator quality from segmentation sensitivity.
(5) We evaluate deployment-aware safeguards and report exploratory subgroup and
downstream EF audits.

\section{Related Work}

\noindent\textbf{Fairness, shortcuts, and privileged information.}
Fairness in medical AI is commonly studied through demographic disparities in
sex, race, and healthcare access
\cite{obermeyer2019dissecting,larrazabal2020gender,seyyedkalantari2021underdiagnosis}.
Benchmark design, metadata availability, workflow assumptions, and metric
choice can also mislead. Models may exploit correlations that work in curated
data but fail at deployment \cite{geirhos2020shortcut}, while model selection
can reward the wrong signal \cite{varoquaux2022machine}.
Conditioning-availability bias is one evaluation-protocol manifestation:
privileged conditioning is rewarded at evaluation. Learning with privileged
information asks whether extra training information helps
\cite{vapnik2009new,lopezpaz2016unifying}; our audit asks whether that gain
survives replacement by a deployable estimate.

\noindent\textbf{Deployment shift and conditional segmentation.}
Scanner/protocol and broader methodological shifts can impair medical-image
models \cite{karani2018lifelong,varoquaux2022machine}. Missing-modality
robustness asks whether a model can operate when an input is absent, whereas
distribution-shift studies ask whether performance transfers to changed data.
Our setting keeps the image and segmentation network fixed and changes only the
quality of the auxiliary variable supplied at evaluation. This isolates whether
a benchmark uses a cleaner conditioning pathway than deployment and directly
covers FiLM-style conditioning \cite{perez2018film}. CAMUS
\cite{leclerc2019camus} and EchoNet-Dynamic \cite{ouyang2020echonet} provide the
echocardiographic test bed. nnU-Net and EchoNet's DeepLabV3--ResNet50 segmenter
serve as non-conditioned architecture references
\cite{isensee2021nnunet,ouyang2020echonet}.

\section{Method}

\subsection{Conditioning-Availability Audit}

Let $X$ be the image, $Y$ the reference segmentation, $C^\ast$ the oracle
conditioning signal, and $\hat C=h(X)$ its deployable estimate. A conditional
model predicts $f_\theta(X,C^\ast)$ under oracle evaluation, but only
$f_\theta(X,\hat C)$ at deployment. A benchmark exhibits
\emph{conditioning-availability bias} when reported performance uses the oracle
path even though clinical inference must use an estimated, missing, or
lower-quality substitute.

For group $g$ with grouping variable $G$, availability indicator $A$, estimation
distance $d$, and segmentation error $\ell$, the three components are
\begin{align}
r_A(g)&=P(A=1\mid G=g), \label{eq:availability}\\
r_E(g)&=\mathbb{E}[d(C^\ast,\hat C)\mid G=g], \label{eq:estimation}\\
r_S(g)&=\mathbb{E}\!\left[\ell(f_\theta(X,\hat C),Y)-
\ell(f_\theta(X,C^\ast),Y)\mid G=g\right]. \label{eq:sensitivity}
\end{align}
These terms separate signal availability, estimation quality, and the extra
segmentation error induced by replacing oracle conditioning. We directly audit
$r_S$, indirectly probe $r_E$, and do not instrument the missing-$C$ regime in
$r_A$.

The processed sequence covers the observed systolic arc. We set ED to
$\varphi=0$ and ES to $\varphi_{\mathrm{ES}}=0.4$; intermediate frames are
linearly positioned on this interval. Each conditioned model is evaluated in
three modes. \textbf{Oracle} supplies this pseudo-phase. \textbf{Estimated} uses
the image-derived phase described below. \textbf{Random} supplies an incorrect
phase drawn uniformly from $[0,0.4]$. We define the gap pair
\begin{equation}
  \Doe=\mathrm{Dice}_{\mathrm{oracle}}-\mathrm{Dice}_{\mathrm{est}},\qquad
  \Dor=\mathrm{Dice}_{\mathrm{oracle}}-\mathrm{Dice}_{\mathrm{rand}}.
  \label{eq:gap}
\end{equation}
The gaps serve different purposes. \Doe estimates deployable-path loss; \Dor
probes sensitivity rather than expected clinical degradation because real phase
errors are not uniform. A large \Doe identifies estimated-path failure; small
\Doe with large \Dor identifies latent sensitivity not triggered by the current
estimator.

\subsection{Model, Training, and Mitigation}

We use a four-level U-Net \cite{ronneberger2015unet} with GroupNorm
\cite{wu2018groupnorm}, base channels 32, bottleneck 512, and $256{\times}256$
inputs. The scalar phase is sinusoidally encoded, passed through an MLP, and used
to FiLM-modulate \cite{perez2018film} all four encoder blocks, the bottleneck,
and all four decoder blocks. Zero-initialized FiLM projections make the initial
segmentation path equivalent to the unconditioned backbone.

The phase head applies global average pooling to bottleneck features and predicts
$s=(\sin 2\pi\varphi,\cos 2\pi\varphi)$. Estimated inference is performed per
frame in two passes. Pass 1 supplies $\varphi=0$ to the FiLM pathway and produces
$s$; the initial value is a fixed design choice that was not ablated. We recover
$\hat\varphi=[\operatorname{atan2}(s_1,s_2)/(2\pi)]\bmod 1$ and clamp it to the
trained systolic interval $[0,0.4]$. Pass 2 recomputes the segmentation with
$\hat\varphi$. Oracle and estimated Dice use the same checkpoint, image,
preprocessing, and segmentation network; only the FiLM input differs.

Training uses Dice--cross-entropy, phase-prediction, adjacent-frame, and
cross-patient cyclic losses. The cyclic term aligns predictions at matched phase
($|\Delta\varphi|\leq0.05$), and losses are introduced in stages.
\emph{Phase collapse} denotes high oracle Dice with sharply degraded estimated
Dice. Strong cyclic consistency can entangle segmentation with exact phase, and
oracle-Dice selection can reward that dependence. S00 is therefore a protocol
falsification test, not a proposed deployment architecture.

We test estimated-Dice checkpoint selection and training-time Gaussian phase
noise/dropout, analogous to domain randomization and conditioning dropout
\cite{tobin2017domain,ho2022classifierfree}. Figure~\ref{fig:concept} summarizes
the inference pathways.

\section{Experiments}

\noindent\textbf{Datasets.}
CAMUS contains 500 apical-four-chamber echocardiograms with four segmentation
classes, split 400/50/50, and provides sex, age, and image-quality metadata
\cite{leclerc2019camus}. EchoNet-Dynamic contains 10,030 apical-four-chamber
videos with LV masks, split 7,465/1,288/1,277, and EF annotations
\cite{ouyang2020echonet}. Checkpoints are selected only on validation data. The
headline audit is then evaluated once on held-out cases: CAMUS $n=50$ and
EchoNet $n=1{,}276$ from the official 1,277-case test split. Study
\texttt{0X5DD5283AC43CCDD1} had no accessible record and was excluded by a
predefined preprocessing rule before model evaluation; every EchoNet model uses
the same 1,276 cases. Subgroup and EF analyses are also held-out audits, and no
hyperparameter is tuned on subgroup, EF, or test outcomes.

\noindent\textbf{Implementation.}
Conditioned models use AdamW \cite{loshchilov2019adamw}, learning rate
$10^{-3}$, weight decay $10^{-4}$, a cosine schedule, effective batch size 8,
100 epochs, and early-stopping patience 40. $\lambda_{\rm phase}=0.1$ throughout. Phase prediction starts at
epoch 21; cyclic consistency starts at epoch 41. These values were fixed across
the sweep.

\noindent\textbf{Model configurations.}
\textbf{B2} is the deployable, unconditioned U-Net with adjacent-frame
consistency. \textbf{P1} uses FiLM conditioning and the phase head but no
adjacent or cyclic loss. \textbf{S00} adds adjacent consistency and strong
cyclic regularization ($\lambda_{\rm cyc}=0.10$), uses no phase perturbation or
EMA, and selects checkpoints by validation oracle Dice; it is an intentional
stress test rather than a proposed deployment model. \textbf{S00$^\prime$} is
identical except for weaker cyclic regularization
($\lambda_{\rm cyc}=0.02$). \textbf{S01} combines the weak cyclic loss with
per-frame Gaussian phase noise ($\sigma=0.05$), phase dropout ($p=0.15$),
estimated-Dice selection, and no EMA. \textbf{P2} uses adjacent and weak cyclic
consistency, coherent per-sequence noise ($\sigma=0.05$) and dropout
($p=0.10$), estimated-Dice selection, a 12-epoch cyclic-loss ramp, and EMA.
\textbf{S05} removes Gaussian noise from P2 while retaining sequence dropout;
\textbf{S08} removes dropout while retaining Gaussian noise; and \textbf{S10}
changes only P2's checkpoint metric from estimated to oracle Dice. Prefix P
denotes a principal conditioned model and S an ablation or safeguard; numeric
suffixes are identifiers, not rankings. nnU-Net and EchoNet's DeepLabV3 (ResNet-50) are non-conditioned
architecture references
\cite{isensee2021nnunet,ouyang2020echonet}.

\begin{figure}
  \centering
  \includegraphics[width=0.98\linewidth]{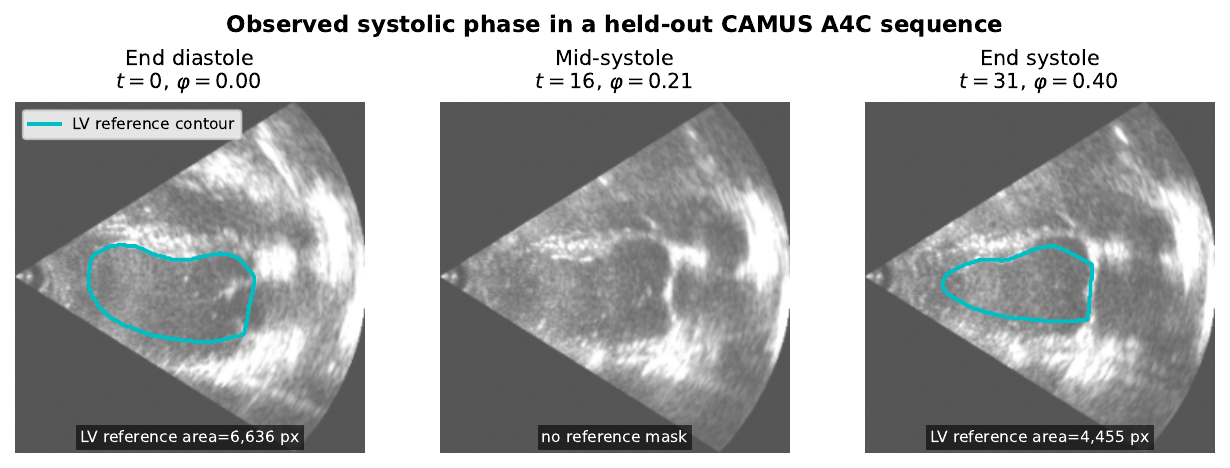}
  \caption{\textbf{Observed systolic phase in A4C echocardiography.}
  A model-independent rule selects the first identifier whose sequence length is
  closest to the test median. The LV contracts from ED ($\varphi=0$) to ES
  ($\varphi=0.4$). CAMUS labels only ED/ES, so the intermediate frame has no
  contour. Deployment must estimate this temporal position from the image.}
  \label{fig:echo_phase}
\end{figure}

\begin{figure}
  \centering
  \includegraphics[width=0.72\linewidth]{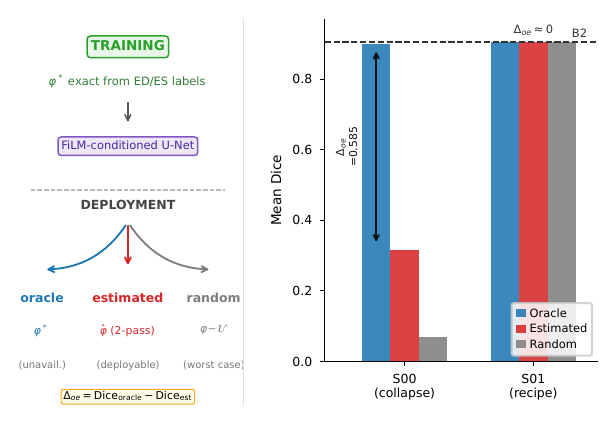}
  \caption{\textbf{Three evaluation pathways.}
  Oracle evaluation supplies the ED/ES-derived phase; estimated evaluation uses
  the two-pass phase head; random evaluation probes dependence on an incorrect
  phase. The right panel reports the validation behavior that motivated the
  held-out audit.}
  \label{fig:concept}
\end{figure}

Figure~\ref{fig:echo_phase} grounds the phase variable in an actual held-out
CAMUS sequence; Figure~\ref{fig:concept} then abstracts the oracle, estimated,
and random conditioning pathways used in the audit.

Each seed yields one validation-selected checkpoint: B2 uses Dice; P1, S01,
and P2 use estimated-mode Dice; S00, S00$^\prime$, and S10 use oracle-mode Dice.
Test, subgroup, and EF outcomes never enter selection. Only S00 has three seeds;
other configurations are single-seed.

\noindent\textbf{Random stress test and uncertainty.}
For each dataset, ten shared random-phase maps use seeds 1000--1009. A phase is
sampled independently for every evaluated ED/ES frame and patient from
$\mathcal U(0,0.4)$, and the same seeded maps are supplied to every conditioned
model. Table~\ref{tab:gap} reports mean random Dice and mean \Dor across maps;
map-level SD quantifies stress-test variability. Table~\ref{tab:gap_ci} reports
approximate 95\% intervals estimated from cohort size and observed dispersion;
these are not patient-bootstrap intervals. Continuous EF outcomes use 10,000
patient-level percentile-bootstrap resamples, while rates use 95\% Wilson
intervals. Seed, patient, and random-map variability are reported separately.

\section{Results}

\subsection{Oracle and Deployable Pathways Diverge}

Table~\ref{tab:gap} separates two behaviors. On CAMUS test, S00 retains oracle
Dice of 0.906 but falls to 0.264 with estimated phase. On EchoNet test, S00-Echo
changes little under its current estimate ($\Doe=0.005$) but degrades under the
random stress test ($\Dor=0.762$). Estimated-only reporting would miss the latter
sensitivity.

Mitigated models remain near B2: CAMUS estimated Dice is 0.909 for S01, 0.905
for P2, and 0.906 for B2; P2-Echo and B2-Echo both round to 0.923. Across ten
maps, mean$\pm$SD \Dor is $0.0027\pm0.0010$ (P1),
$0.8684\pm0.0145$ (S00), $0.0021\pm0.0007$ (S01), and
$0.0023\pm0.0005$ (P2) on CAMUS; EchoNet gives $0.0039\pm0.0004$,
$0.7620\pm0.0026$, and $0.0013\pm0.0002$, respectively. All three S00 seeds
retain high oracle Dice and large oracle--random gaps, but severe estimated-path
collapse occurs in one: random-phase sensitivity replicates, while collapse
magnitude is seed-dependent.

\begin{figure}
  \centering
  \includegraphics[width=0.7\linewidth]{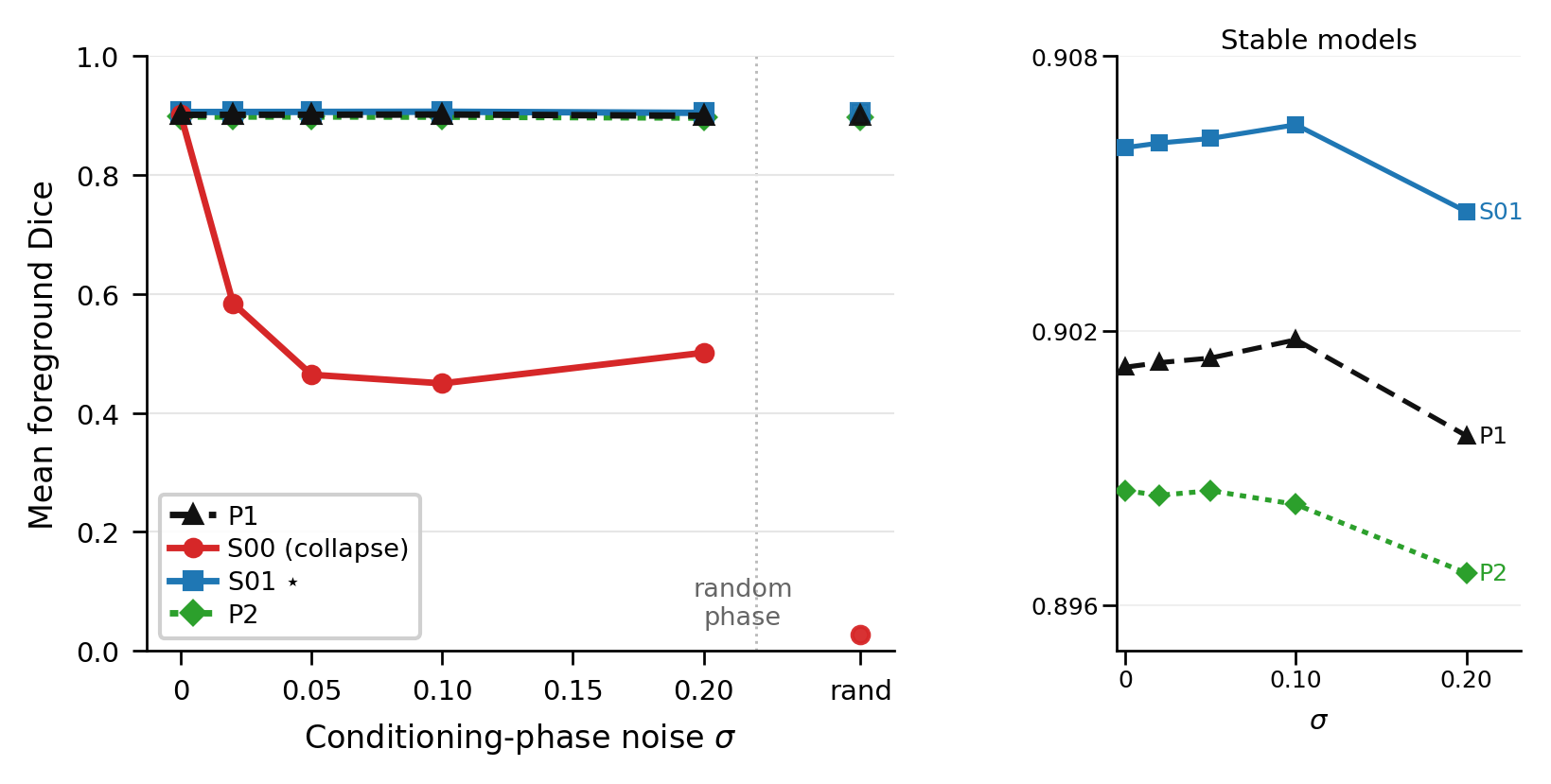}
  \caption{\textbf{Phase-noise robustness.}
  Dice under Gaussian phase noise $\sigma\in\{0,0.02,0.05,0.10,0.20\}$ and
  random phase. S00 degrades steeply; P1, S01, and P2 remain flat, showing that \Dor
  reflects continuous sensitivity rather than one random draw.}
  \label{fig:noise}
\end{figure}

Figure~\ref{fig:noise} further shows that this is not a single random-draw
effect: S00 degrades continuously as phase noise increases, whereas P1, S01, and P2
remain stable.

\begin{table}
  \caption{Held-out test conditioning-availability audit. Dice is mean foreground
  Dice. Random Dice and \Dor are means over ten shared random maps; uncertainty intervals and map-level standard deviations are
  reported in Table~\ref{tab:gap_ci}. $\star$ denotes the primary
  CAMUS mitigation selected by validation estimated-mode Dice, not statistical
  significance. nnU-Net and DeepLabV3 (ResNet-50) are non-conditioned
  architecture references, and B2 is the deployable unconditioned reference.}
  \label{tab:gap}
  \centering
  \fontsize{8}{9}\selectfont
  \setlength{\tabcolsep}{3.0pt}
  \begin{tabular}{llcccccc}
    \toprule
    Dataset & Model & Oracle & Est. & Random & \Doe & \Dor & Interpretation \\
    \midrule
    \multirow{6}{*}{CAMUS test}
      & nnU-Net & -- & 0.916 & -- & -- & -- & published ref. \\
      & B2 ref. & -- & 0.906 & -- & -- & -- & deployable ref. \\
      & P1 & 0.906 & 0.899 & 0.903 & 0.007 & 0.003 & small gaps \\
      & S00 & 0.906 & 0.264 & 0.038 & 0.642 & 0.868 & est. failure \\
      & S01 ($\star$) & 0.912 & 0.909 & 0.910 & 0.003 & 0.002 & small gaps \\
      & P2 & 0.908 & 0.905 & 0.906 & 0.003 & 0.002 & small gaps \\
    \midrule
    \multirow{5}{*}{EchoNet test}
      & DeepLabV3-R50 & -- & 0.924 & -- & -- & -- & published ref. \\
      & B2-Echo ref. & -- & 0.923 & -- & -- & -- & deployable ref. \\
      & P1-Echo & 0.924 & 0.918 & 0.920 & 0.006 & 0.004 & small gaps \\
      & S00-Echo & 0.922 & 0.917 & 0.160 & 0.005 & 0.762 & random-sensitive \\
      & P2-Echo & 0.923 & 0.923 & 0.922 & 0.001 & 0.001 & small gaps \\
    \bottomrule
  \end{tabular}
\end{table}

\begin{table}
  \caption{Approximate 95\% intervals for the principal held-out audit.
  Oracle, estimated, and paired-gap intervals use cohort size and observed
  dispersion rather than patient bootstrap; map SD uses ten shared phase maps.}
  \label{tab:gap_ci}
  \centering
  \fontsize{7.5}{8.5}\selectfont
  \setlength{\tabcolsep}{2.2pt}
  \begin{tabular}{llccccc}
    \toprule
    Dataset & Model & Oracle interval & Est. interval & \Doe interval
      & \Dor interval & Map SD \\
    \midrule
    \multirow{4}{*}{CAMUS test}
      & P1  & [0.897, 0.914] & [0.889, 0.909]
      & [0.002, 0.013] & [$-0.001$, 0.007] & 0.0010 \\
      & S00 & [0.897, 0.914] & [0.225, 0.303]
      & [0.602, 0.681] & [0.856, 0.879] & 0.0145 \\
      & S01 ($\star$) & [0.905, 0.919] & [0.902, 0.917]
      & [0.001, 0.006] & [$-0.001$, 0.005] & 0.0007 \\
      & P2  & [0.900, 0.916] & [0.896, 0.913]
      & [0.001, 0.006] & [$-0.001$, 0.005] & 0.0005 \\
    \midrule
    \multirow{3}{*}{EchoNet test}
      & P1-Echo  & [0.922, 0.926] & [0.916, 0.920]
      & [0.005, 0.007] & [0.003, 0.005] & 0.0004 \\
      & S00-Echo & [0.920, 0.924] & [0.915, 0.919]
      & [0.004, 0.006] & [0.754, 0.770] & 0.0026 \\
      & P2-Echo  & [0.921, 0.925] & [0.921, 0.925]
      & [0.000, 0.002] & [0.000, 0.002] & 0.0002 \\
    \bottomrule
  \end{tabular}
\end{table}

\subsection{Mechanism and Safeguards}

Table~\ref{tab:mechanism} isolates the CAMUS mechanism. Reducing cyclic weight
changes S00's estimated Dice from 0.316 to 0.894. Estimated-mode selection and
phase perturbation then reduce the remaining gap. These are validation ablations;
differences among the non-collapsing configurations are descriptive single-seed
comparisons.

\begin{table}
  \caption{CAMUS validation ablation; pf = per-frame perturbation, ps =
  per-sequence perturbation, Sel. = checkpoint selection. $\star$ denotes the
  primary mitigation chosen by validation estimated-mode Dice, not statistical
  significance. Bold marks the highest estimated Dice and smallest \Doe among
  conditioned configurations at the displayed precision.}
  \label{tab:mechanism}
  \centering
  \fontsize{8}{9}\selectfont
  \setlength{\tabcolsep}{3.1pt}
  \begin{tabular}{llcccccc}
    \toprule
    Config & Change tested & $\lambda_{\rm cyc}$ & Pert. & Sel. & EMA
    & Est. Dice & \Doe \\
    \midrule
    P1 & Phase only & 0 & -- & est. & \no & 0.897 & 0.004 \\
    S00$^\prime$ & Weak cyclic ctrl & 0.02 & none & oracle & \no & 0.894 & 0.007 \\
    S00 & Strong cyclic & 0.10 & none & oracle & \no & 0.316 & 0.585 \\
    \midrule
    S01 ($\star$) & Full CAMUS recipe & 0.02 & pf & est. & \no
      & \textbf{0.906} & \textbf{$<0.001$} \\
    S05 & No Gaussian noise & 0.02 & ps & est. & \yes & 0.905 & 0.003 \\
    P2 & ps + EMA & 0.02 & ps & est. & \yes & 0.897 & 0.002 \\
    S08 & No dropout & 0.02 & ps & est. & \yes & 0.901 & 0.004 \\
    S10 & Oracle selection & 0.02 & ps & oracle & \yes & 0.893 & 0.004 \\
    \midrule
    B2 ref. & Unconditioned & -- & -- & -- & -- & 0.906 & -- \\
    \bottomrule
  \end{tabular}
\end{table}

Table~\ref{tab:phase} shows that estimator quality does not explain collapse.
P1 and S00 have nearly identical phase MAE (0.194 vs.\ 0.195 over 1,600 frames),
yet \Doe differs by two orders of magnitude (0.004 vs.\ 0.585). The failure is
therefore driven by segmenter hypersensitivity to conditioning error rather
than a substantially worse phase estimator. S01 and P2 retain small gaps even
though their estimators remain imperfect, further separating the quality of the
phase head from the sensitivity of the segmentation pathway.

\begin{table}
  \caption{CAMUS validation phase-estimator characterization over 1,600 frames.
  Phase MAE is circular mean absolute error. Similar estimator quality but
  divergent \Doe isolates segmenter sensitivity from estimator error.}
  \label{tab:phase}
  \centering
  \fontsize{8}{9}\selectfont
  \setlength{\tabcolsep}{7pt}
  \begin{tabular}{lccc}
    \toprule
    Model & Phase MAE$\downarrow$ & \Doe & \Dor \\
    \midrule
    P1  & 0.194 & 0.004 & 0.004 \\
    S00 & 0.195 & 0.585 & 0.832 \\
    S01 & 0.144 & $<0.001$ & $<0.001$ \\
    P2  & 0.165 & 0.002 & 0.005 \\
    \bottomrule
  \end{tabular}
\end{table}

\subsection{Held-Out Subgroup and EF Audits}

Table~\ref{tab:subgroup} audits available CAMUS test strata. S00 has a large
gap throughout, but small strata preclude ranking harm or claiming significant
between-stratum differences. The audit measures $r_S$, indirectly reflects the
current $r_E$ pathway, and does not measure $r_A$, site access, or
protected-attribute exposure. Results are exploratory deployment-severity
estimates, not causal demographic unfairness. The same ten phase maps reproduce
the overall estimate up to rounding; S01 reduces all displayed gaps near zero.

\begin{table}
  \caption{Held-out CAMUS test subgroup deployment-severity audit. Values are
  patient-level Dice gaps. Strata are descriptive point estimates; poor-quality
  is especially underpowered because $n=5$.}
  \label{tab:subgroup}
  \centering
  \fontsize{8}{9}\selectfont
  \setlength{\tabcolsep}{2.8pt}
  \begin{tabular}{lccccc}
    \toprule
    Subgroup & $n$ & S00 \Doe & S00 \Dor & S01 \Doe & S01 \Dor \\
    \midrule
    All & 50 & 0.642 & 0.868 & 0.003 & 0.002 \\
    Female & 12 & 0.625 & 0.862 & 0.001 & 0.001 \\
    Male & 38 & 0.648 & 0.870 & 0.003 & 0.003 \\
    Young (28--56 yr) & 17 & 0.622 & 0.882 & 0.002 & 0.003 \\
    Mid (57--72 yr) & 16 & 0.691 & 0.829 & 0.002 & 0.001 \\
    Older ($\geq$73 yr) & 17 & 0.617 & 0.890 & 0.004 & 0.004 \\
    Good quality & 28 & 0.634 & 0.901 & 0.001 & 0.001 \\
    Medium quality & 17 & 0.666 & 0.824 & 0.005 & 0.005 \\
    Poor quality & 5 & 0.608 & 0.828 & 0.002 & 0.002 \\
    \bottomrule
  \end{tabular}
\end{table}

EchoNet test EF-stratification gives a larger-cohort check. Across Normal
($n=991$), Reduced ($n=161$), and Severely reduced EF ($n=124$), P2-Echo differs
from B2-Echo by only $-0.0011$, $+0.0006$, and $+0.0005$ Dice, respectively.
Thus closing the availability gap does not reveal an EF-group penalty on
held-out EchoNet cases.

Table~\ref{tab:clinical} propagates estimated-mode CAMUS masks into EF using
oracle ED/ES frames and the GT-reference single-plane Simpson method. Empty,
degenerate, non-finite, or EF values outside $[-100,100]$ are invalid. S00 fails
every case by the Dice threshold, leaves 28\% invalid EF, and has large EF-MAE.
S01 and P2 remove segmentation failures; P2 removes invalid EF, while S01 leaves
one. Their EF-MAE intervals overlap B2 and signed biases remain negative. Thus,
closing the segmentation gap prevents degenerate masks but does not establish
better EF measurement.

\section{Discussion and Conclusion}

\noindent\textbf{Deployment-validity blind spot.}
S00 shows why oracle validation is unsafe: oracle Dice can match strong
baselines while deployment fails or remains brittle, and the divergence persists
on unseen patients. FDA's AI/ML action plan emphasizes lifecycle oversight and
real-world performance monitoring \cite{fda2021aiml}; the oracle--estimated gap
directly checks the inference pathway.

\noindent\textbf{Why three pathways matter.}
CAMUS S00 has estimated-path collapse, with both \Doe and \Dor large; EchoNet
S00-Echo has phase brittleness, with small \Doe but large \Dor. Oracle-only
evaluation misses the former, whereas estimated-only reporting misses the
latter. The three-pathway evaluation separates deployment loss from latent
conditioning dependence.

\noindent\textbf{Fairness relevance and limitations.}
Conditioning-availability bias is primarily a deployment-validity failure: a
benchmark may certify a pathway that some sites, scanners, patients, or
workflows do not receive. It becomes fairness-relevant when conditioning is
differentially available or estimated with different error across sites,
scanners, operators, acquisition quality, or patient strata. We do not claim
that cardiac phase is a protected attribute or that subgroup gaps establish
causal demographic unfairness. The experiments directly audit $r_S$,
indirectly probe $r_E$ through the current estimator, and do not measure $r_A$.
CAMUS subgroup findings remain exploratory because strata are small. The EF
audit uses GT-mask EF from a single-plane Simpson reference rather than
prospective clinical labels. We also do not test fully missing $C$, prospective
workflow availability, or cross-site phase-estimator error. Future work should
define application-specific \Doe thresholds, audit prospective cohorts, and
measure real conditioning availability.

\begin{table}
  \caption{Held-out CAMUS test downstream EF audit in estimated mode ($n=50$).
  EF-MAE and EF-bias are against GT-mask EF from the same single-plane Simpson
  method. Failure = Dice $<0.70$; EF invalid = degenerate or nonphysiological EF.
  Dice, EF-MAE, and EF-bias intervals are 10,000-resample patient bootstrap
  intervals; rate intervals are Wilson 95\% intervals. EF valid $n$ is the
  denominator for EF-MAE and EF-bias.}
  \label{tab:clinical}
  \centering
  \fontsize{8}{9}\selectfont
  \setlength{\tabcolsep}{2pt}
  \begin{tabular}{@{}l
    >{\centering\arraybackslash}p{0.16\linewidth}
    >{\centering\arraybackslash}p{0.12\linewidth}
    >{\centering\arraybackslash}p{0.07\linewidth}
    >{\centering\arraybackslash}p{0.12\linewidth}
    >{\centering\arraybackslash}p{0.15\linewidth}
    >{\centering\arraybackslash}p{0.16\linewidth}@{}}
    \toprule
    Model & Dice$_{\rm est}$ [95\% CI] & Failure\% [95\% CI]
      & EF valid $n$ & EF invalid\% [95\% CI]
      & EF-MAE [95\% CI] & EF-bias [95\% CI] \\
    \midrule
    B2 ref. & 0.906 [0.898, 0.914] & 0 [0, 7.1] & 50
      & 0 [0, 7.1] & 8.4 [6.6, 10.2] & $-2.6$ [$-5.4$, 0.5] \\
    P1  & 0.899 [0.889, 0.909] & 0 [0, 7.1] & 50
      & 0 [0, 7.1] & 10.7 [8.5, 13.0] & $-8.0$ [$-11.0$, $-4.9$] \\
    S00 & 0.264 [0.225, 0.303] & 100 [92.9, 100] & 36
      & 28 [17.5, 41.7] & 39.2 [30.2, 48.8] & $-4.1$ [$-20.5$, 11.3] \\
    S01 ($\star$) & 0.909 [0.902, 0.917] & 0 [0, 7.1] & 49
      & 2 [0.4, 10.5] & 9.4 [6.6, 12.8] & $-6.3$ [$-10.2$, $-2.8$] \\
    P2  & 0.905 [0.896, 0.913] & 0 [0, 7.1] & 50
      & 0 [0, 7.1] & 9.2 [6.9, 11.9] & $-3.9$ [$-7.1$, $-0.3$] \\
    \bottomrule
  \end{tabular}
\end{table}

\noindent\textbf{Conclusion.}
We formalize conditioning-availability bias as a deployment-validity failure.
The held-out gaps expose estimated-path collapse and latent phase brittleness.
A strong-cyclic, oracle-selected configuration collapses in one run and remains
random-phase sensitive across three; deployment-aware selection and perturbation
close the evaluated segmentation gaps. Preventing degenerate masks does not by
itself improve EF error or bias. Conditional models should receive credit only
for gains that survive the deployed pathway and downstream audit.

\begin{credits}
\subsubsection{\ackname} Computational resources for model training were provided
by the VinUni-Illinois Smart Health Center (VISHC) server infrastructure. This research was funded by the National Foundation for Science and Technology Development (NAFOSTED) through Project No. IZVSZ2\_229539 (2025–2027).
\end{credits}

\bibliographystyle{splncs04}
\bibliography{references}

\end{document}